\documentclass[10pt,twocolumn,letterpaper]{article}

\usepackage{cvpr}              
\usepackage{stfloats}
\usepackage{graphicx}
\usepackage{subcaption}
\usepackage{algorithm}
\usepackage{algorithmic}
\usepackage{amsmath}
\usepackage{amssymb}
\usepackage{microtype}
\usepackage{booktabs}
\usepackage{array}
\usepackage{marvosym}
\usepackage{xcolor}
\usepackage{colortbl}
\usepackage{array}
\usepackage{tabularx}
\usepackage{ragged2e}
\usepackage{float}
\usepackage{cuted}
\usepackage{multirow}
\usepackage{bm}

\newcolumntype{L}[1]{>{\raggedright\arraybackslash}p{#1}}

\definecolor{cvprblue}{rgb}{0.21,0.49,0.74}
\usepackage[pagebackref,breaklinks,colorlinks,allcolors=cvprblue]{hyperref}

\def\paperID{} 
\def\confName{CVPR}
\def\confYear{2026}

\title{CLaD: Planning with Grounded Foresight via Cross-Modal Latent Dynamics}

\author{
    Andrew Jeong \quad Jaemin Kim \quad Sebin Lee \quad Sung-Eui Yoon \\
    \\KAIST
    \vspace{-1cm}
}

\begin{document}
\maketitle
\begin{strip}
    \centering
    \vspace{-0.2cm}
    \includegraphics[width=0.98\textwidth]{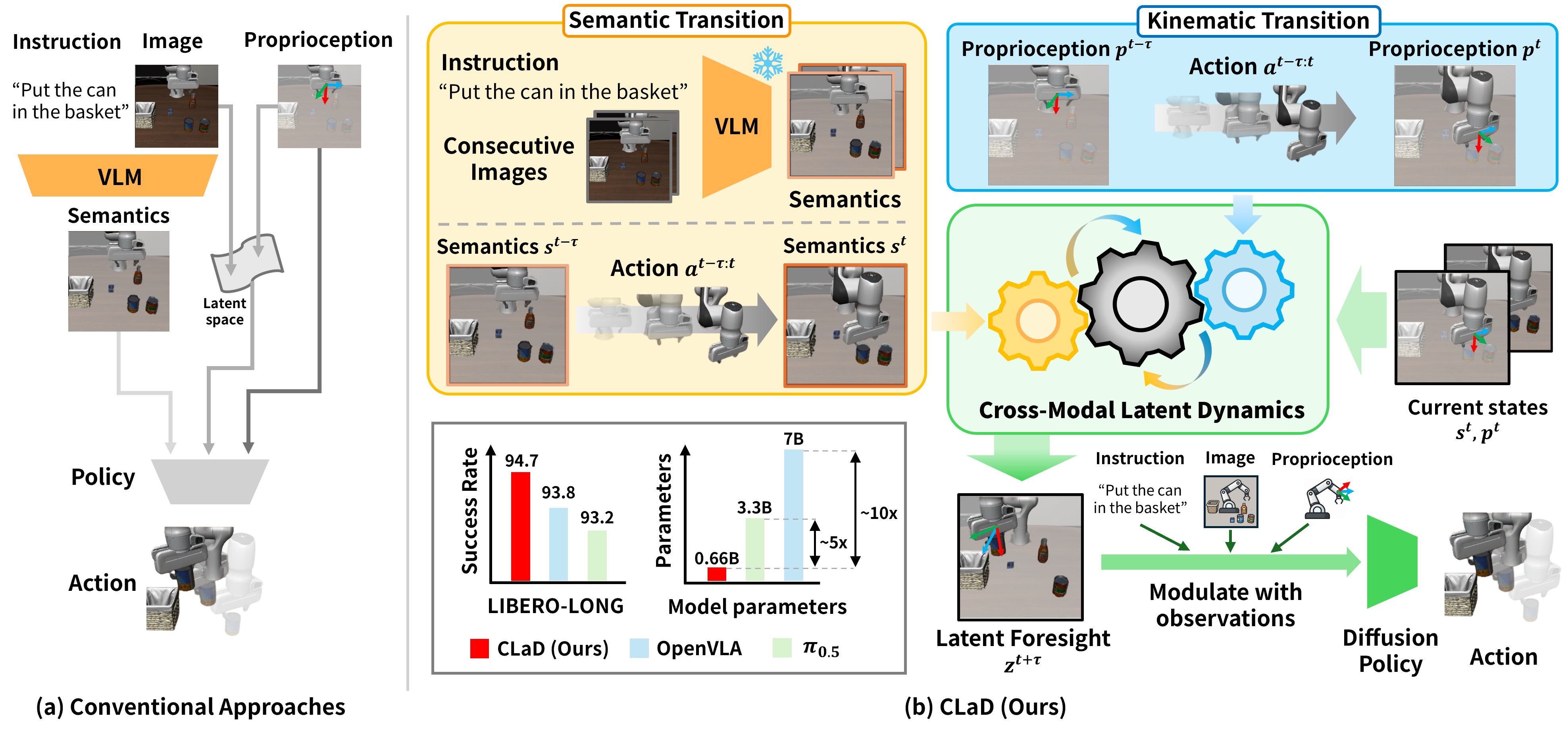}
    \captionof{figure}{\textbf{Overview of CLaD.} (a) Conventional approaches either generate semantic artifacts (e.g., subgoal images or texts), or plan in unimodal latent spaces that lack cross-modal understanding. (b) CLaD learns cross-modal latent dynamics to predict grounded latent foresights, which condition a diffusion policy for action generation. CLaD achieves 94.7\% with only 0.66B parameters, competitive with OpenVLA (7B) and $\pi_{0.5}$ (3.3B).}
    \label{fig:overview}
    \vspace{-0.2cm}
\end{strip}
\begin{abstract}
Robotic manipulation involves kinematic and semantic transitions that are inherently coupled via underlying actions. However, existing approaches plan within either semantic or latent space without explicitly aligning these cross-modal transitions. To address this, we propose CLaD, a framework that models how proprioceptive and semantic states jointly evolve under actions through asymmetric cross-attention that allows kinematic transitions to query semantic ones. CLaD predicts grounded latent foresights via self-supervised objectives with EMA target encoders and auxiliary reconstruction losses, preventing representation collapse while anchoring predictions to observable states. Predicted foresights are modulated with observations to condition a diffusion policy for action generation. On LIBERO-LONG benchmark, CLaD achieves 94.7\% success rate, competitive with large VLAs with significantly fewer parameters.\\
Project page: \small{\url{https://andrewwwj.github.io/clad}}
\end{abstract}
\vspace{-0.45cm}
\section{Introduction}
\label{sec:intro}
When a robot grasps an object, two coupled changes occur: proprioceptive transitions (robot moving) and semantic transitions (visual scene changing). This coupling suggests that robots should jointly reason about what they see and how they move to accomplish manipulation tasks. While recent robot learning methods have shown significant advancements, effectively coordinating proprioception and semantics during planning remains underexplored. Recent planning frameworks either generate computationally expensive semantic artifacts during planning ~\cite{black2024susie,zawalski2024robotic, wen2025diffusionvla}, or plan within latent state spaces~\cite{hafner2025mastering, hansen2024tdmpc2}, which are more efficient but lack explicit constraints about how different modalities are correlated during a transition. In the absence of such constraints, the latent representations of semantic and kinematic states may decouple during rollout, potentially leading to trajectories that are physically or logically inconsistent.

We suggest \textit{Cross-modal Latent Dynamics} (CLaD), which models how both proprioceptive and semantic states jointly evolve under actions. Unlike prior approaches that align static states across modalities~\cite{radosavovic2023real, ze2024gnfactor}, our key insight is that consistency should be enforced over \textit{transitions}: when a robot closes its gripper (kinematic transition), it watches how the scene changes (semantic transition), and these co-occurring changes are causally coupled through the underlying action.  We formalize this coupling via asymmetric cross-attention, where proprioceptive transitions serve as queries to semantic ones, enabling the model to interpret scene-level changes through the robot's kinematic context.

CLaD adopts a two-stage framework. In the first stage, we capture cross-modal dynamics from semantic and proprioceptive transitions and learn \textit{grounded latent foresights} from the dynamics, via self-supervised objectives combining EMA-based latent prediction with auxiliary reconstruction constraints. In the second stage, a diffusion policy is conditioned on the foresights through observation-based FiLM modulation. This separation allows planning to focus on accurate future state prediction without a bias about policy optimization ~\cite{hafner2025mastering}, while operating entirely within a compact latent space to avoid the computational overhead of explicit semantic generation.

\vspace{-10pt}
\paragraph{Core Contributions:}
\begin{itemize}[leftmargin=*,topsep=2pt,itemsep=1pt]
\item \textbf{Cross-Modal Latent Dynamics.} We propose a cross-modal dynamics model that learns how proprioceptive and semantic transitions jointly evolve under actions via asymmetric cross-attention, interpreting a semantic transition through a proprioceptive transition cue.
\item \textbf{Grounded Latent Foresight.} We introduce a two-stage training framework that predicts compact future latent states from the shared cross-modal dynamics, which are grounded to observable quantities, enabling them to serve as subgoals for diffusion-based control.
\item \textbf{Parameter-Efficient Planning.} CLaD achieves competitive performance to OpenVLA (7B) and $\pi_{0.5}$ (3.3B) on LIBERO-LONG with significantly fewer parameters (0.66B), demonstrating that grounded latent foresight enables efficient and scalable robot planning within a compact latent space.
\end{itemize}

\section{Preliminaries}
\label{sec:preliminary}

\subsection{Diffusion Model}

Diffusion models represent generative models that have demonstrated remarkable success in various domains, which is theoretically grounded in Langevin Dynamics \cite{vincent2011connection, song2019generative}. Among various approaches, Denoising Diffusion Probabilistic Model (DDPM) \cite{ho2020denoising, song2021score} is a representative framework for current diffusion models.

In DDPM, the forward diffusion process adds Gaussian noise to data $\mathbf{x}_0$ over $T$ timesteps through a Markov chain:
\begin{equation}
q(\mathbf{x}_t | \mathbf{x}_0) = \mathcal{N}(\mathbf{x}_t; \sqrt{\bar{\alpha}_t}\mathbf{x}_0, (1-\bar{\alpha}_t)\mathbf{I}),
\end{equation}
where $\bar{\alpha}_t = \prod_{i=1}^{t} \alpha_i$ and $\alpha_t \in (0, 1)$ is a variance schedule controlling the noise level at timestep $t$.

The model is trained to predict the noise $\boldsymbol{\epsilon}$ added during the forward process by minimizing:
\begin{equation}
\mathcal{L}_{\text{Diffusion}} = \mathbb{E}_{t, \mathbf{x}_0, \boldsymbol{\epsilon}} \left[ \|\boldsymbol{\epsilon} - \boldsymbol{\epsilon}_\theta(\mathbf{x}_t, t)\|^2 \right],
\end{equation}
where $\boldsymbol{\epsilon} \sim \mathcal{N}(\mathbf{0}, \mathbf{I})$ is the noise added to obtain $\mathbf{x}_t$ from $\mathbf{x}_0$. At inference, sampling starts from $\mathbf{x}_T \sim \mathcal{N}(\mathbf{0}, \mathbf{I})$ and iteratively denoises to generate a data sample:
\begin{equation}
\mathbf{x}_{t-1} = \frac{1}{\sqrt{\alpha_t}}\left(\mathbf{x}_t - \frac{1-\alpha_t}{\sqrt{1-\bar{\alpha}_t}}\boldsymbol{\epsilon}_\theta(\mathbf{x}_t, t)\right) + \sigma_t \mathbf{z},
\end{equation}
where $\mathbf{z} \sim \mathcal{N}(\mathbf{0}, \mathbf{I})$ and $\sigma_t$ is the sampling noise schedule.

\subsection{Diffusion Policy}

Diffusion models naturally extend to sequential decision-making by conditioning the denoising process on observations. Diffusion Policy \cite{chi2023diffpolicy} applies DDPM to generate actions conditioned on visual or proprioceptive observations. The model learns to denoise an action sequence $\mathbf{a}_{0:H}$ by the noise prediction network $\boldsymbol{\epsilon}_\theta(\mathbf{a}_k, k, \mathbf{o}_k)$ conditioned on current observations $\mathbf{o}_k$. This formulation enables the policy to generate smooth, multi-modal action distributions while avoiding covariate shift issues common in autoregressive policies. The training objective remains:
\begin{equation}
\mathcal{L}_{\text{DP}} = \mathbb{E}_{\mathbf{a}_0, k, \boldsymbol{\epsilon}, \mathbf{o}} \left[ \|\boldsymbol{\epsilon} - \boldsymbol{\epsilon}_\theta(\mathbf{a}_k, k, \mathbf{o}_k)\|^2 \right],
\end{equation}
where the $\mathbf{a}_k$ is a noisy action obtained from the action sample $\mathbf{a}_0$. At execution time, the policy samples action chunks by running the reverse diffusion process conditioned on the current observation, enabling expressive and robust visuomotor control.

\section{Related Work}
\label{sec:related_work}

\subsection{Planning with Semantic Reasoning}

Recent semantic reasoning approaches for robotic manipulation can be categorized into text-based reasoning and visual generation methods. Text-based approaches leverage language models for planning, where SayCan~\cite{ahn2022saycan} grounds language in affordances for feasible action selection, while chain-of-thought methods~\cite{zawalski2024robotic} generate explicit reasoning steps to decompose tasks and improve interpretability. Recent unified architectures~\cite{wen2025diffusionvla} combine autoregressive textual reasoning with diffusion-based action generation to address the limitations of pure autoregressive or diffusion models. Image generation methods use predicted images or videos as planning mechanisms, including subgoal image generation~\cite{black2024susie} or text-conditioned video generation~\cite{du2023learning, tian2025seer, wu2024unleashing} to guide low-level policies . Despite significant progress, these methods face limitations about computational overhead due to iterative reasoning through generating semantic artifacts.

\subsection{Planning in Latent Space}

Latent-space planning achieves computational efficiency by operating in compact learned representations. World model-based planning, which learns forward dynamics for model-predictive control, either by explicitly imagining rollouts in recurrent state-space models~\cite{hafner2025mastering} or by implicitly optimizing trajectories in a decoder-free latent space~\cite{hansen2024tdmpc2}. Planning as latent inference, which treats planning as probabilistic inference over latent trajectory distributions, using techniques like posterior inference~\cite{rosete2023taco} or backward reasoning from the goal~\cite{lbp}. Hierarchical methods learn multi-scale dynamics where slow higher-level latents provide context for fast lower-level dynamics~\cite{gumbsch2024thick, hansen2025hierarchical}, while compositional approaches decompose tasks into reusable primitives~\cite{he2024pivot,zhou2024robodreamer}. Despite their efficiency, existing methods learn latent representations that implicitly conflate semantic and kinematic information without explicit mechanisms to ensure different modalities evolve consistently.

\subsection{Cross-Modal Representation in Robotics}

Prior work on cross-modal representation learning in robotics has primarily focused on aligning visual, linguistic, and proprioceptive modalities at the level of individual observations. Language-visual alignment methods learn such correspondences through contrastive objectives, aligning visual observations with language instructions~\cite{li2024decisionnce, nair2022r3m}. A complementary line of work addresses proprioceptive-visual alignment, jointly embedding robot kinematic states and image observations into a shared representation space via heterogeneous pre-trained transformers~\cite{jiang2025robots, wang2024scaling}. However, these methods model cross-modal correspondences at individual timesteps, without capturing how semantic and kinematic states jointly change under actions. Consequently, they provide no explicit mechanism to ensure that changes in the visual scene remain consistent with changes in the robot's joint configurations.

To address this, we introduce a cross-modal latent dynamics model that learns correlations between \textit{transitions} rather than static states. By modeling the co-evolution of semantic and proprioceptive states under actions, our framework predicts latent foresight from a cross-modal dynamics that interprets a semantic transition through a kinematic transition, followed by a diffusion policy to perform expressive low-level control given the foresight.

\section{Planning with Cross-Modal \! Latent \! Dynamics}
\label{sec:method}

Consider a human playing tennis: as the arm extends toward the ball (proprioceptive change), the visual scene updates with the racket approaching the ball (semantic change). These transitions in different modalities are coupled through the underlying action. We formalize this intuition by learning \textit{cross-modal latent dynamics} that capture how proprioceptive and semantic states  jointly evolve under actions.

In Section~\ref{sec:clad}, ~\ref{sec:foresight}, we introduce \textit{CLaD}, a cross-modal latent dynamics model that learns to predict future latent states, or latent foresights, by modeling a correlation between transitions of different modalities (System 2~\cite{stanovich2000individual, kahneman2011thinking}). In Section~\ref{sec:diffusion_policy}, a diffusion policy performs low-level control generating action sequences conditioned on these predicted foresights (System 1~\cite{stanovich2000individual, kahneman2011thinking}).

\subsection{Cross-modal Latent Dynamics}
\label{sec:clad}

Prior cross-modal learning approaches~\cite{radosavovic2023real, ze2024gnfactor} align states across modalities, matching visual features with proprioceptive states at single timesteps. However, we propose shifting from learning correlations between static states to learning correlations between dynamic \textit{transitions}. In other words, we focus on understanding how modalities change together: when a robot moves its joints (kinematic transition), the distance between the robot and object becomes closer (semantic transition). Our objective is to learn a representation that captures how such heterogeneous transitions relate with respect to their common dynamic context.

\begin{figure*}[!t]
    \centering
    \includegraphics[width=\textwidth]{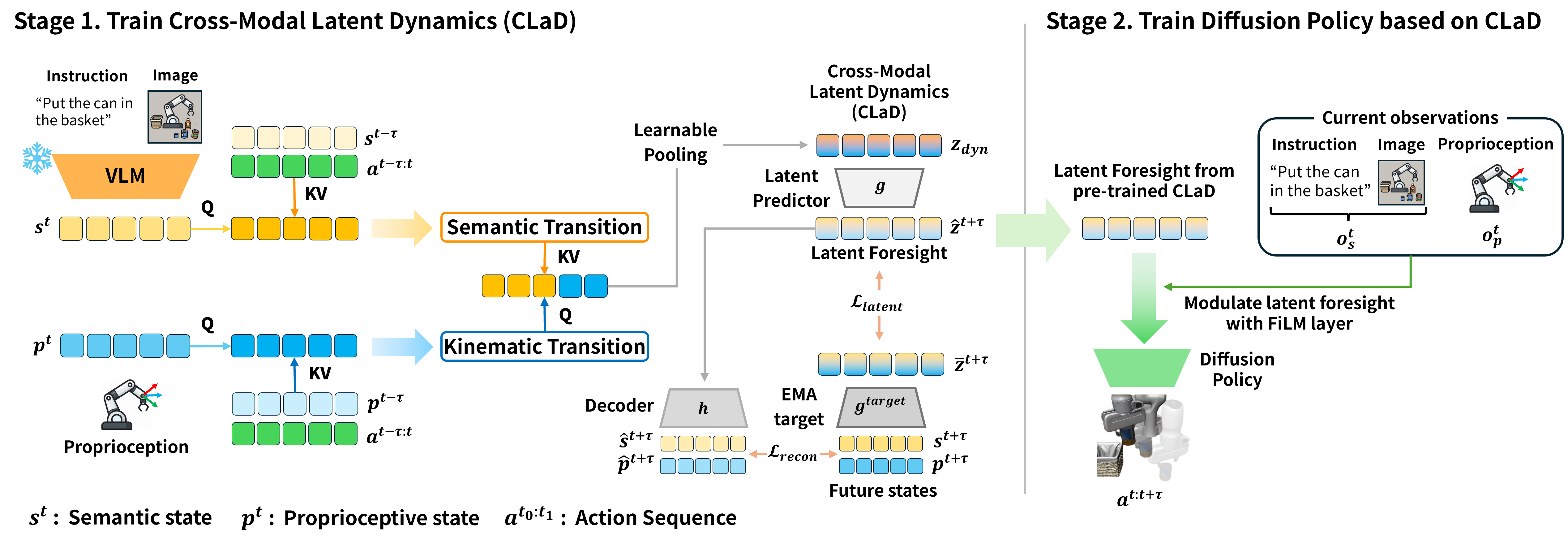}
    \caption{\textbf{Detailed architecture of CLaD's two-stage framework.} 
        Stage 1 (Left) - Semantic state $s_t$ and proprioceptive state $p_t$ are each encoded and fed into modality-specific cross-attention modules to extract transition representations $z_s$ and $z_p$. Asymmetric cross-attention, where proprioceptive transitions query semantic ones, yields shared dynamics $z_{\text{dyn}}$, from which
        lightweight MLPs predict future latent foresights $\hat{z}^{t+\tau}_{p}$ and $\hat{z}^{t+\tau}_{s}$. These predictions are supervised via EMA target encoders using actual
        future states $s_{t+\tau}$ and $p_{t+\tau}$, with auxiliary reconstruction decoders maintaining correspondence with observable quantities and preventing representation collapse. 
        Stage 2 (Right) - Predicted latent foresights $\hat{z}^{t+\tau} = [\hat{z}^{t+\tau}_{p};\, \hat{z}^{t+\tau}_{s}]$ are adaptively modulated with current observations via FiLM layers, and the resulting representation conditions a diffusion policy to generate action sequences $a_{t:t+\tau}$.}
    \label{fig:architecture}
    \vspace{-0.2cm}
\end{figure*}

\paragraph{Transition embedding.}
Formally, in an MDP, a transition is characterized by a probability of next state given current state and an action. As shown in Figure~\ref{fig:architecture} (left), at timestep $t$, we encode a proprioceptive state $p^t \in \mathbb{R}^{D_p}$ containing joint angles and velocities, and a semantic state $s^t=\text{FiLM}(v^t, l) \in \mathbb{R}^{D_s}$ where FiLM layer~\cite{perez2018film} fuses vision-language embeddings $v^t, l$ from a frozen pre-trained VLM~\cite{li2024decisionnce}:
\begin{align}
\mathbf{p}^t &= f_p(p^t) \in \mathbb{R}^{N_p \times H}, \\
\mathbf{s}^t &= f_s(s^t) \in \mathbb{R}^{N_s \times H},
\end{align}
where $f_p\in\mathbb{R}^{D_p \rightarrow H}$, $f_s\in\mathbb{R}^{D_s \rightarrow H}$ are MLP encoders with hidden dimension $H$ and their outputs are tokenized into sequences of length $N_p$ and $N_s$, respectively.

Since our diffusion policy operates with action horizon $\tau$, we design the dynamics model to predict a future state at this temporal granularity. To facilitate learning cross-modal transition relationships, we extract transition representations for each semantics and proprioception as:
\begin{align}
\mathbf{z}_p &= CrossAttn(\mathbf{p}^t, [\mathbf{p}^{t-\tau}; \mathbf{a}^{t-\tau:t}]) \in \mathbb{R}^{N_p \times H}, \label{eq:zp} \\
\mathbf{z}_s &= CrossAttn(\mathbf{s}^t, [\mathbf{s}^{t-\tau}; 
\mathbf{a}^{t-\tau:t}]) \in \mathbb{R}^{N_s \times H}.
\end{align}
where $[;]$ denotes concatenation over token dimension, $\mathbf{p}^{t-\tau}, \mathbf{s}^{t-\tau}, \mathbf{p}^t, \mathbf{s}^t $ denote the past states at $\tau$ steps prior and the current states of both modalities, and $\mathbf{a}^{t-\tau:t} = f_a(a^{t-\tau:t})$ denotes the encoded action sequence. During training, we stochastically replace action tokens with a learnable token, similar to masking strategies in masked autoencoders~\cite{he2022masked}. This encourages the model to infer transitions from state differences to improve robustness over actions. This representation enables the extraction of a \textit{transition embedding} that compress heterogeneous transition structures into compact embeddings to capture the causal dependency of future state on past state-action pairs, which facilitates the extraction of cross-modal dynamic patterns.

\paragraph{Learning cross-modal dynamics.}
To learn correlated transition of both modalities, we employ asymmetric cross-attention where proprioceptive transitions query semantic transitions (see the left side of Figure~\ref{fig:architecture}):
\begin{equation}
\mathbf{z}_{p\rightarrow s} = CrossAttn(\mathbf{z}_p, \mathbf{z}_s) \in \mathbb{R}^{N_p \times H}.
\end{equation}

By using proprioceptive transitions (e.g., changes in robot joint configurations) as queries, the model interprets semantic changes (e.g., object displacement) through the robot's kinematic context. We validate this design against alternative cross-attention configurations in Section~\ref{sec:ablation}.
Finally, we project $\mathbf{z}_{p\rightarrow s}$ to a compact representation:
\begin{equation}
\mathbf{z}_{\text{dyn}} = Pool(\mathbf{q}_{\text{out}}, \mathbf{z}_{p\rightarrow s}) \in \mathbb{R}^{H}.
\end{equation}
where $\it{Pool}$ denotes a learnable pooling~\cite{jaegle2021perceiver} that maps inputs into the learnable latent query $\mathbf{q}_{\text{out}} \in \mathbb{R}^{H}$. Compared to mean or max pooling, this readout structure encourages the model to capture the salient cross-modal dynamics patterns. We call $\mathbf{z}_{\text{dyn}}$ as \textbf{CLaD} (Cross-modal Latent Dynamics).

\subsection{Learning Grounded Latent Foresight}
\label{sec:foresight}

Unlike text-based reasoning~\cite{zawalski2024robotic, wen2025diffusionvla} or visual generation methods~\cite{black2024susie} that require expensive iterative generation of semantic artifacts (e.g., subgoal images or texts), our approach predicts latent foresights from the learned cross-modal latent dynamics $\mathbf{z}_{\text{dyn}}$. These foresights serve as implicit future states that guide action generation without the computational overhead of explicit semantic reasoning.

\paragraph{Learning objective for grounded latent foresight.}
We predict latent foresights from the learned cross-modal dynamics $\mathbf{z}_{\text{dyn}}$ using a self-supervised learning scheme in latent space. This formulation enables the model to learn abstract yet task-relevant representations of future states grounded in both semantic and proprioceptive contexts.
Specifically, we predict future embeddings, which we term \textit{latent foresight}, $\hat{\mathbf{z}}^{t+\tau}$ for both modalities using lightweight MLP decoders $g_p: \mathbb{R}^H \rightarrow \mathbb{R}^H$ and $g_s: \mathbb{R}^H \rightarrow \mathbb{R}^H$ given cross-modal dynamics $\mathbf{z}_{\text{dyn}}$:
\begin{align}
\hat{\mathbf{z}}_p^{t+\tau} &= g_p(\mathbf{z}_{\text{dyn}}) \in \mathbb{R}^{H}, \label{eq:pred_p} \\
\hat{\mathbf{z}}_s^{t+\tau} &= g_s(\mathbf{z}_{\text{dyn}}) \in \mathbb{R}^{H}, \label{eq:pred_s} \\
\hat{\mathbf{z}}^{t+\tau} &= [\hat{\mathbf{z}}_p^{t+\tau}; \hat{\mathbf{z}}_s^{t+\tau}]
\label{eq:concat_foresight}
\end{align}
where both predictions condition solely on shared dynamics $\mathbf{z}_{\text{dyn}}$. This design ensures predicted future states respect the cross-modal dynamics.

A critical challenge in latent space prediction is representation collapse~\cite{chen2020simple}. Following established self-supervised learning (SSL) methods~\cite{grill2020bootstrap, caron2021emerging, assran2023self, bardes2024revisiting}, we leverage exponential moving average (EMA) target encoders $f_p^{\text{target}}, f_s^{\text{target}}, \text{FiLM}^{\text{target}}$ updated with momentum $m$ following standard practice~\cite{grill2020bootstrap}:
\begin{equation}
\theta^{\text{target}} \leftarrow m \cdot \theta^{\text{target}} + (1 - m) \cdot \theta. \label{eq:ema}
\end{equation}
These slowly-updating targets provide stable learning signals, preventing the online encoder from chasing a moving target. Target embeddings are computed as:
\begin{align}
\bar{\mathbf{z}}_p^{t+\tau} &= f_p^{\text{target}}(\mathbf{p}^{t+\tau}), \label{eq:target_p}\\
\bar{\mathbf{z}}_s^{t+\tau} &= f_s^{\text{target}}(\mathbf{s}^{t+\tau}). \label{eq:target_s}
\end{align}
where $\mathbf{p}^{t+\tau}$ denotes a future proprioceptive state with $\tau$ steps ahead and $\mathbf{s}^{t+\tau}=\text{FiLM}^{\text{target}}(\mathbf{v}^{t+\tau}, \mathbf{l})$ denotes a future semantics state from next image after $\tau$ steps and language instruction, where both states are encoded with corresponding EMA target encoders.

Referring to recent SSL work~\cite{assran2023self, oquab2023dinov2}, we use MSE on $L_2$-normalized embeddings, constraining embeddings to a unit hypersphere to prevent magnitude collapse while preserving angular relationships between embeddings that encode semantic similarity between prediction and target:
\begin{align}
\mathcal{L}_{\text{latent}} = \left\| \hat{\mathbf{z}}_p^{t+\tau} - \frac{\bar{\mathbf{z}}_p^{t+\tau}}{\|\bar{\mathbf{z}}_p^{t+\tau}\|} \right\|_2^2 + \left\| \hat{\mathbf{z}}_s^{t+\tau} - \frac{\bar{\mathbf{z}}_s^{t+\tau}}{\|\bar{\mathbf{z}}_s^{t+\tau}\|} \right\|_2^2. \label{eq:latent_loss}
\end{align}

In order to ensure latent representations remain anchored in observable quantities, we add auxiliary reconstruction losses using lightweight decoders $h_p\in \mathbb{R}^{H \rightarrow D_p}, h_s\in \mathbb{R}^{H \rightarrow D_s}$:
\begin{equation}
\mathcal{L}_{\text{recon}} = \| h_p(\hat{\mathbf{z}}_p^{t+\tau}) - \mathbf{p}^{t+\tau} \|_1 + \| h_s(\hat{\mathbf{z}}_s^{t+\tau}) - \mathbf{s}_v^{t+\tau} \|_1. \label{eq:recon_loss}
\end{equation}

While latent prediction induces a semantically structured embedding hypersphere, the reconstruction loss anchors the representations to observable quantities. Together, these complementary objectives preserve task-relevant semantics while preventing the latent space from drifting toward excessive abstraction, ensuring that the predicted latent foresights remain grounded in observable contexts.

The combined objective is:
\begin{equation}
\mathcal{L} = \mathcal{L}_{\text{latent}} + \lambda_{\text{recon}} \mathcal{L}_{\text{recon}}, \label{eq:total_loss}
\end{equation}
where $\lambda_{\text{recon}} = 0.1$ balances latent prediction with reconstruction, giving higher weight to latent representation learning.

We term our predictions as a \textit{grounded} latent foresight as it is anchored through two mechanisms above: (1) prediction targets are computed from observed future states via EMA target encoders~\cite{grill2020bootstrap} and (2) auxiliary reconstruction losses $\mathcal{L}_{\text{recon}}$ constrain latent embeddings to decode to raw proprioceptive and visual observations, maintaining predictions to remain grounded to each states of both modalities.

\subsection{Diffusion Policy Guided by Latent Foresight}
\label{sec:diffusion_policy}

Standard diffusion policies~\cite{chi2023diffpolicy} generate action sequences directly from current observations, \ie, $p(a_t \mid o_t)$, without access to anticipated future states. CLaD extends this by conditioning action generation on predicted latent foresights $z^t$, yielding $p(a_t \mid o_t, z^t)$. This formulation is structurally analogous to goal-conditioned policies~\cite{black2024susie, lbp}, where $z^t$ serves as a learned latent subgoal rather than an explicitly generated observation, avoiding the computational overhead of iterative semantic generation~\cite{zawalski2024robotic, wen2025diffusionvla, zhao2025cot}.

\paragraph{Foresight-conditioned diffusion control.}
Given predicted latent foresights $z^t_p, z^t_s$ from the frozen CLaD, we first encode the current proprioceptive and semantic observations using modality-specific encoders $e_p$ and $e_s$:

\begin{align}
\mathbf{o}_p^t = e_p(\mathbf{p}^t), \quad \mathbf{o}_s^t = e_s(\mathbf{s}_v^t, \mathbf{s}_l).
\end{align}

We then condition each foresight embedding on its corresponding current observation via FiLM modulation~\cite{perez2018film}, where the observation supplies the affine scale and shift parameters that anchor the predicted foresight to the present observational context:

\begin{align}
\mathbf{g}_p = \text{FiLM}(\hat{\mathbf{z}}^{t+\tau}, \mathbf{o}_p^t),\quad 
\mathbf{g}_s = \text{FiLM}(\hat{\mathbf{z}}^{t+\tau}, \mathbf{o}_s^t).
\end{align}

The policy is trained with the standard DDPM noise-prediction objective~\cite{ho2020denoising}, along with observation-modulated foresight $\mathbf{g}_p$, $\mathbf{g}_s$:
\begin{align}
    \mathcal{L}_{\text{policy}}
        = \mathbb{E}_{\mathbf{a}_0,\, k,\, \epsilon}
          \!\left[\,
            \bigl\|\,
              \epsilon
              - \hat{\epsilon}_\theta(\mathbf{a}_k,\, k,\, \mathbf{g}_p, \mathbf{g}_s)
            \,\bigr\|^2_2
          \,\right],
    \label{eq:policy_loss}
\end{align}

where $\mathbf{a}_0 \sim q(\mathbf{a}_0)$ is the ground-truth action sequence, $\mathbf{a}_k = \sqrt{\bar{\alpha}_k}\mathbf{a}_0 + \sqrt{1-\bar{\alpha}_k}\boldsymbol{\epsilon}$ is the noised action at diffusion step $k$, and $\boldsymbol{\epsilon} \sim \mathcal{N}(0, \mathbf{I})$.
\section{Experiments}
\label{sec:exp}

\begin{table*}[t]
\centering
\caption{\textbf{Performance comparison on LIBERO-LONG benchmark}. CLaD achieves competitive performance with significantly fewer parameters (0.66B) compared to baselines. LIBERO-LONG tasks include: (1) put soup and sauce in basket; (2) put box and butter in basket; (3) turn on stove and put pot; (4) put bowl in drawer and close it; (5) put mugs on left and right plates; (6) pick book and place it in back; (7) put mug on plate and put pudding to right; (8) put soup and box in basket; (9) put both pots on stove; (10) put mug in microwave and close it.}
\begin{tabular}{cc@{\hspace{1cm}}c@{\hspace{1cm}}c@{\hspace{1cm}}c@{\hspace{1cm}}c@{\hspace{1cm}}c@{\hspace{1cm}}c@{\hspace{1cm}}c}
\toprule
\multirow{2}{*}{Task ID} & Susie$^\dagger$ & Seer$^\dagger$ & LBP$^\dagger$ & ${\pi _0}^\ddagger$& ${\pi _{0.5}}^\ddagger$ & OpenVLA$^\ddagger$ & Ours$^\dagger$ & Ours$^\ddagger$ \\
 & (0.86B) & (0.32B) & (0.19B) & (3.3B) & (3.3B) & (7B) & \multicolumn{2}{c}{(0.66B)} \\
\midrule
1 & 83.3 & 88.3 & 90.0 & 74.0 & 92.0 & 98.0 & 100.0& 100.0\\
2 & 63.3 & 90.0 & 100.0 & 94.0 & 98.0 & 96.0 & 100.0& 100.0\\
3 & 96.6 & 98.3 & 100.0 & 88.0 & 98.0 & 100.0 & 98.3  & 98.0\\
4 & 100.0 & 100.0 & 100.0 & 22.0 & 98.0 & 78.0 & 93.7  & 94.0\\
5 & 83.3 & 91.7 & 76.6 & 100.0 & 100.0 & 84.0 & 92.0  & 91.0\\
6 & 83.3 & 93.3 & 86.6 & 88.0 & 94.0 & 100.0 & 100.0 & 100.0\\
7 & 83.3 & 85.0 & 90.0 & 98.0 & 96.0 & 94.0 & 94.3  & 95.0\\
8 & 39.9 & 91.7 & 86.6 & 86.0 & 100.0 & 96.0 & 100.0 & 100.0\\
9 & 53.3 & 61.7 & 60.0 & 76.0 & 62.0 & 92.0 & 81.7  & 82.0\\
10 & 76.6 & 71.7 & 96.6 & 94.0 & 94.0 & 100.0 & 85.0  & 87.0\\
\midrule
\textbf{Avg. SR}
   & \textbf{76.3}
   & \textbf{87.7}
   & \textbf{88.6}
   & \textbf{82.0}
   & \textbf{93.2}
   & \textbf{93.8}
   & \textbf{94.5}
   & \textbf{94.7} \\
\bottomrule
\end{tabular}

\vspace{2mm}
\raggedright\footnotesize
$\dagger$: \textbf{Avg. SR} is defined as an average over 20 rollouts from the top 3 checkpoints. For other baselines, the results are reported in the respective papers.\\ 
$\ddagger$: \textbf{Avg. SR} is defined as an average over 50 rollouts. The results are reported in LIBERO-PRO~\cite{liberopro} as primary papers report only the average success rate. 
\label{tab:main_results}
\end{table*}

We evaluate CLaD on LIBERO-LONG, a benchmark requiring multi-step sequential reasoning and precise control across 10 long-horizon manipulation tasks. Our experiments validate three core hypotheses: (1) modeling cross-modal latent dynamics enables parameter-efficient long-horizon planning demonstrating competitive performance with larger VLA models, (2) robust latent foresight requires both semantic and proprioceptive modalities to be jointly grounded rather than used in isolation, and (3) asymmetric cross-attention that proprioceptive transition queries semantic transition provides the most effective inductive bias for extracting task-relevant dynamics.

\subsection{Experimental Setup}
\label{sec:setup}

\paragraph{Implementation details.}

CLaD uses hidden dimension $H=1024$ with learnable tokens $N_p, N_s=4$ and action horizon $\tau=6$. Stage 1 is trained for 25K steps with batch size 128, EMA momentum $m = 0.995$, $\lambda_{\text{recon}}=0.1$, requiring 2 hours, and Stage 2 is trained for 200K steps with batch size 128, requiring 20 hours. Total model size is 0.66B parameters (VLM: 0.1B, CLaD: 0.33B, Policy: 0.23B) where we utilize DecisionNCE~\cite{li2024decisionnce} as VLM. All experiments are done with NVIDIA RTX 4090 GPU.

We compare against recent vision-language-action models across diverse scales and planning paradigms. We report available published results and follow each method's evaluation protocol. All methods are evaluated on the same LIBERO-LONG benchmark to ensure fair comparison.

\textbf{SuSIE}~\cite{black2024susie} uses an image-editing diffusion model as a high-level planner that proposes visual subgoals, followed by a low-level policy executes actions to reach each subgoal.

\textbf{Seer}~\cite{tian2025seer} proposes predictive inverse dynamics model that predict robot actions via inverse dynamics conditioned on forecasted future visual states.

\textbf{LBP}~\cite{lbp} performs latent space backward planning by grounding the task into a final latent goal and then recursively predicting intermediate subgoals toward the current state.

\textbf{OpenVLA}~\cite{kim2024openvla} fine-tunes a Llama~2-based VLM on 970k Open X-Embodiment demonstrations, serving as a strong open-source 7B-parameter generalist VLA baseline.

$\bm{\pi_0}$~\cite{black2024pi_0} and $\bm{\pi_{0.5}}$~\cite{pi0.5} are composed of a VLM backbone with a flow-matching action expert, pre-trained on large-scale robot demonstrations for multi-task robotic control.

\paragraph{Dataset and tasks.}
We use LIBERO-LONG~\cite{liu2023libero}, which contains 10 long-horizon manipulation tasks in kitchen and tabletop environments, where each task requires 2-3 sequential subtasks. We follow the standard training protocol~\cite{kim2024openvla} and average the success rates of top-3 checkpoints over 20 rollouts for methods marked as $\dagger$, or evaluate over 50 rollouts with a single checkpoint marked as $\ddagger$.

\subsection{Results}
\label{sec:main_results}
\paragraph{Performance on long-horizon planning}

Table~\ref{tab:main_results} shows that CLaD achieves a 94.7\% average success rate, surpassing OpenVLA~\cite{kim2024openvla} (93.8\%, 7B) and $\pi_{0.5}$~\cite{pi0.5} (93.2\%, 3.3B) with only 0.66B parameters. CLaD exhibits relatively stable performance across the 10 LIBERO-LONG tasks compared to other baselines, suggesting that grounded cross-modal foresight contributes to consistent behavior over extended horizons. Among similar-scale methods, CLaD surpasses SuSIE~\cite{black2024susie} (76.3\%, 0.86B) by 18.2 points and Seer~\cite{tian2025seer} (87.7\%, 0.32B) by 6.8 points, highlighting the benefit of explicit cross-modal dynamics relative to generation-based semantic planning approaches. LBP~\cite{lbp} is also built upon a latent-space planning framework and achieves solid average performance with fewer parameters (88.6\%, 0.19B), but it shows pronounced degradation in certain tasks, especially on Task~9 (put both pots on the stove) which involves perceptually similar objects and fine-grained alignment, while CLaD achieves better performance with 81.3\% success rate compared to LBP. It suggests that planning with explicit cross-modal dynamics helps resolve perceptual ambiguity.

\paragraph{Computational Efficiency}

Table~\ref{tab:computation_cost} and Table~\ref{tab:planning_time} compare CLaD's computational efficiency against baseline methods. CLaD requires fewer parameters (0.66B) and less memory (4~GB) due to its latent planning framework, compared to OpenVLA~\cite{kim2024openvla} (7B, 15~GB) and $\pi_{0.5}$~\cite{pi0.5}(3.3B, 19~GB). Table~\ref{tab:planning_time} further compares CLaD against comparable latent planning methods in terms of planning efficiency. Although CLaD uses more parameters than LBP~\cite{lbp} (0.19B) and UVA~\cite{li2025unified} (0.5B), it achieves a higher success rate of 94.7\% (+6.1\% over LBP, +4.7\% over UVA), while maintaining a planning latency of 0.012~s that enables real-time deployment. These results demonstrate that the performance gain of CLaD justifies the modest parameter increase over other latent planners.

\begin{table}[h]
    \centering
    \caption{\textbf{Comparison on computational requirements.} CLaD achieves faster inference and lower resource consumption compared to large-scale VLA models, while maintaining competitive task performance on LIBERO-LONG.}
    \label{tab:computation_cost}
    \begin{tabularx}{\linewidth}{c >{\centering\arraybackslash}X >{\centering\arraybackslash}X >{\centering\arraybackslash}X}
    \toprule
    \textbf{Method} & OpenVLA & $\pi_{0.5}$ & \textbf{CLaD} \\
    \midrule
    \textbf{Inference time (Hz)} & 6 & 10 & \textbf{25}\\
    \textbf{Memory (GB)} & 15  & 19 & \textbf{4}\\
    \textbf{Parameters (B)} & 7.0 & 3.3 & \textbf{0.66}\\
    \midrule
    \end{tabularx}
\end{table}

\begin{table}[h]
    \centering
    \caption{\textbf{Performance and efficiency of latent planning methods on LIBERO-LONG.} Although CLaD requires more parameters than LBP (0.19B) and UVA (0.5B), it yields higher average success rate, respectively, suggesting a competitive trade-off between model capacity and task performance within the latent planning paradigm.}
    \label{tab:planning_time}
    \begin{tabularx}{\linewidth}{c >{\centering\arraybackslash}X >{\centering\arraybackslash}X >{\centering\arraybackslash}X}
    \toprule
    \textbf{Method} & UVA & LBP & \textbf{CLaD}\\
    \midrule
    \textbf{Parameters (B)} & 0.5 & 0.19 & \textbf{0.66} \\
    \textbf{Planning Time (s)} & 0.195 & 0.008 & \textbf{0.012}\\
    \textbf{Avg. SR (\%)}  & 90.0 & 88.6 & \textbf{94.7} \\
    \bottomrule
    \end{tabularx}
\end{table}

\begin{figure*}[!t]
    \centering
    \label{fig:modality_dropout}
    \includegraphics[width=\textwidth]{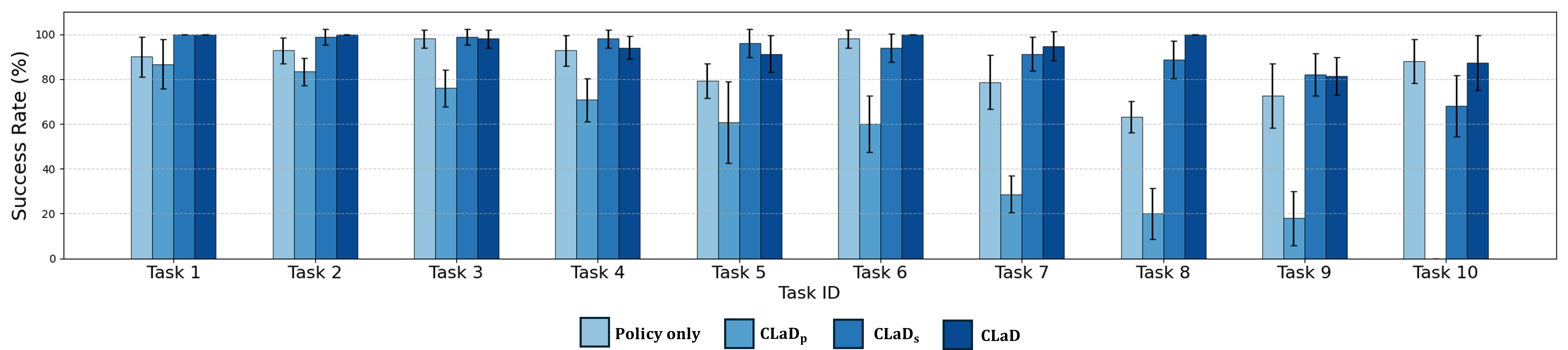}
    \caption{\textbf{Modality contribution analysis on LIBERO-LONG tasks}. We compare diffusion policy without foresight conditioning (Policy only), conditioning on proprioceptive foresight only ($\text{CLaD}_\textbf{p}$), semantic foresight only ($\text{CLaD}_\textbf{s}$), and full cross-modal foresight (CLaD). Results demonstrate that cross-modal foresight provides substantial gains over single-modality predictions, though semantic foresight alone achieves reasonable performance. Notably, proprioceptive foresight alone severely degrades performance, suggesting that kinematic predictions require semantic context for effective planning.}
    \vspace{-0.4cm}
\end{figure*}

\subsection{Ablation Studies}
\label{sec:ablation}

\paragraph{Modality contribution analysis.} We ablate the contribution of each modality by comparing four variants: policy without foresight conditioning, proprioceptive foresight only, semantic foresight only, and full cross-modal foresight. Semantic foresight alone ($\text{CLaD}_\textbf{s}$: 91.5\%) improves over the no-foresight baseline (Policy only: 84.8\%), confirming that the semantic future state provides meaningful guidance for long-horizon manipulation. Notably, proprioceptive foresight alone ($\text{CLaD}_\textbf{p}$: 50.4\%) substantially degrades performance far below the baseline, indicating that kinematic predictions without semantic grounding introduce misleading conditioning signals. The additional performance gain from $\text{CLaD}_\textbf{s}$ (91.5\%) to CLaD (94.7\%) further demonstrates that jointly modeling how kinematic and visual transitions co-evolve yields better results than single-modality foresight.

\vspace{-0.3cm}
\paragraph{Effect of $\mathcal{L}_{\text{recon}}$ in grounding latent foresight}

\begin{table}[!b]
    \centering
    \vspace{-0.2cm}
    \caption{\textbf{Ablation of the auxiliary reconstruction loss $\mathcal{L}_\text{recon}$.} Removing $\mathcal{L}_\text{recon}$ leads to 8.6\% performance degradation, indicating that $\mathcal{L}_\text{recon}$ is critical to ground latent dynamics $z_{\text{dyn}}$ to observable states for learning robust latent foresights, not merely a regularizer.}
    \renewcommand{\arraystretch}{1.3}
    \begin{tabular}{ccc}
    \hline
    \textbf{Learning objective} & \textbf{$\mathcal{L}_{\text{latent}}$} & \textbf{$\mathcal{L}_{\text{latent}}$ + $\mathcal{L}_{\text{recon}}$} \\
    \hline
    \textbf{Avg. SR} & 86.1 \textbf{(-8.6)} & 94.7 \\
    \hline
    \end{tabular}
    \label{tab:aux_loss}
    \vspace{0.45cm}
\end{table}

\begin{figure}[!t]
    \centering
    \includegraphics[width=\linewidth]{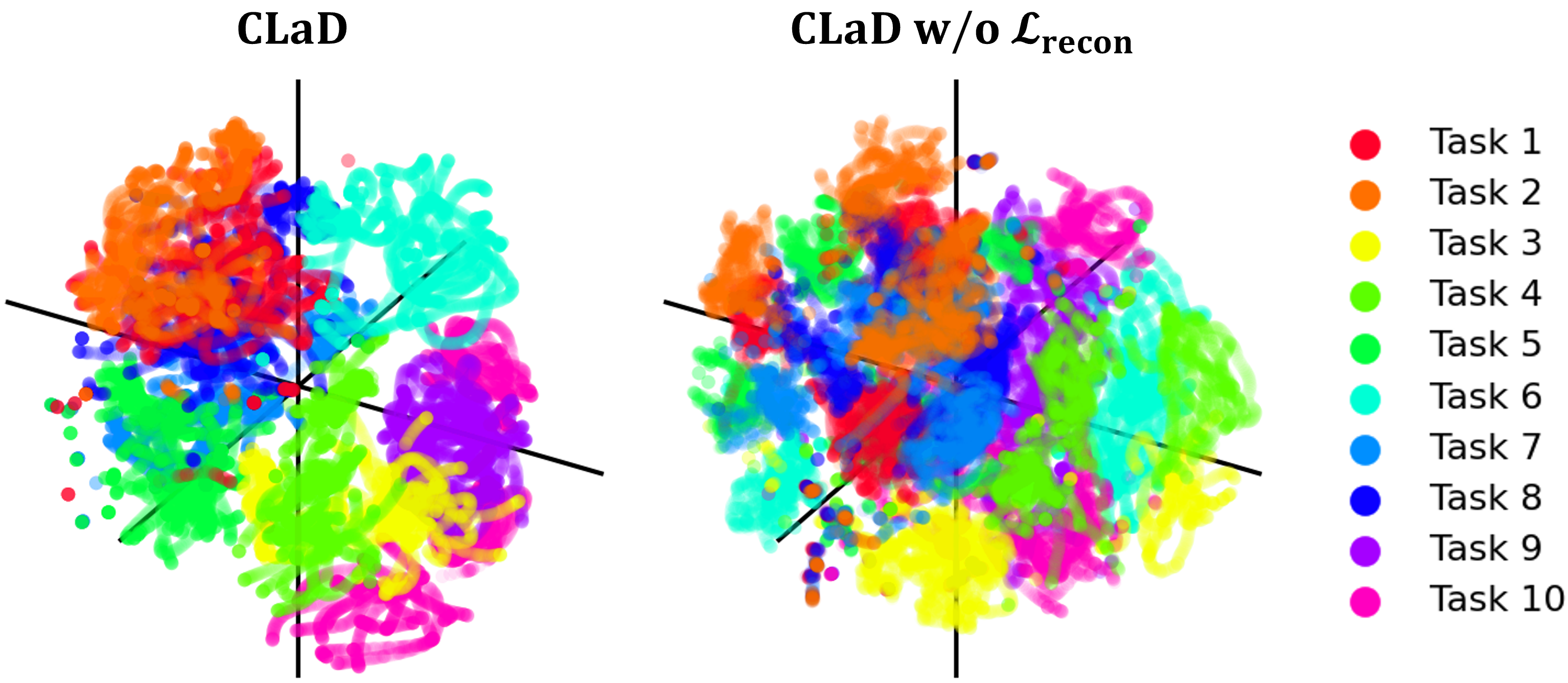}
    \caption{\textbf{UMAP of latent embedding $z_{\text{dyn}}$ with regard to $\mathcal{L}_\text{recon}$.} With $\mathcal{L}_\text{recon}$ (left), embeddings form distinct task-specific clusters. Without $\mathcal{L}_{\text{recon}}$ (right), clusters become diffuse and overlapping, indicating reduced semantic structure in the latent space.}
    \label{fig:umap}
    \vspace{-0.2cm}
\end{figure}

Table~\ref{tab:aux_loss} shows that removing $\mathcal{L}_{\text{recon}}$ results in substantially lower success rates (86.1\%) than the baseline (94.7\%) across manipulation tasks. This result indicates that the latent prediction loss $\mathcal{L}_{\text{latent}}$ alone is insufficient to maintain latent representations that reliably condition the diffusion policy. We attribute this degradation to representation drift: without reconstruction supervision, the model is free to optimize $\mathcal{L}_{\text{latent}}$ through increasingly abstract embeddings that match EMA target but lose correspondence with observable quantities.
By requiring predicted latent foresights to remain decodable to their respective proprioceptive and semantic state spaces, $\mathcal{L}_{\text{recon}}$ constrains the latent space against this drift, not merely as an auxiliary regularizer, but as a grounding mechanism for the dynamics representation $z_{\text{dyn}}$. To examine how this grounding manifests in the learned representation structure, we visualize $z_{\text{dyn}}$ using UMAP~\cite{mcinnes2018umap} across the 10 LIBERO-LONG tasks (see Figure~\ref{fig:umap}). When trained with $\mathcal{L}_{\text{recon}}$ (left side of Figure~\ref{fig:umap}), the model produces distinct clusters that align with task contexts: Tasks 1, 2, and 8, which share similar instruction templates (``put both [objects] in the basket'') and visual scene configurations, cluster tightly yet remain separable, while Task 6 characterized by a distinct instruction pattern (``pick up the [object] and place it in the [location]'') and different object layouts, occupies a clearly distinct region of the latent space. In contrast, removing $\mathcal{L}{\text{recon}}$ leads to diffuse, overlapping clusters: the previously separable groupings of Tasks 1, 2, and 8 overlaps each other, indicating that the latent space has lost sensitivity to meaningful distinctions over task contexts. Together, the quantitative results and representation analysis consistently show that $\mathcal{L}_{\text{recon}}$ is essential for learning latent foresights that are both structurally organized and grounded in observable state spaces, leading to effective foresight-conditioned policy learning.

\paragraph{Asymmetric cross-attention design.} Table~\ref{tab:attention_ablation} compares three cross-attention configurations. Symmetric self-attention performs worst (86.7\%), confirming that undirected cross-modal exchange fails to capture the dependency between kinematic and visual transitions. Both asymmetric configurations outperform this baseline, demonstrating that directionality in cross-modal interaction is important for learning the latent dynamics. Using proprioceptive transitions as queries over semantic ones (ours: 94.7\%) yields the best performance, marginally surpassing the reverse direction (93.8\%). This ordering supports the intuition that kinematic transitions provide a more grounded basis for interpreting scene-level visual changes, reflecting a useful inductive bias aligned with robot manipulation.


\begin{table}[h]
    \centering
    \vspace{-0.2cm}
    \caption{\textbf{Ablation of cross-attention configurations.} Using proprioceptive transitions as queries over semantic transitions (94.7\%) outperforms both the reverse configuration (93.8\%) and symmetric self-attention (86.7\%), confirming that directing kinematic context to attend over semantic transitions is a beneficial inductive bias for extracting cross-modal dynamics.}
    \label{tab:attention_ablation}
    \begin{tabular}{lc}
    \toprule
    \textbf{Attention Configuration} & \textbf{Avg. SR(\%)} \\
    \midrule
    Symmetric self-attention & 86.7 \\
    Semantics queries proprioceptive & 93.8 \\
    Proprioceptive queries semantic (ours) & \textbf{94.7} \\
    \bottomrule
    \end{tabular}
    \vspace{-0.3cm}
\end{table}

\section{Conclusion}
\label{sec:conclusion}
We introduced CLaD (Cross-modal Latent Dynamics), a framework that models cross-modal latent dynamics to predict grounded foresight in robotic manipulation. Unlike prior work that aligns static observations across modalities, CLaD learns how proprioceptive and semantic states co-evolve under actions through asymmetric cross-attention, where kinematic transitions query semantic ones to interpret visual scene changes through the robot's kinematic context. Grounded latent foresights are predicted from the shared dynamics representation $z_\text{dyn}$ via EMA-based target predictions to future latent states, while auxiliary reconstruction losses ground the representation to observable states. A diffusion policy is conditioned on these foresights to generate desirable actions. On LIBERO-LONG benchmark, CLaD achieves 94.7\% average success rate, attaining competitive performance with large VLAs with much fewer parameters.

\section{Limitations and future work.} While CLaD demonstrates effective performance on long-horizon tasks with few parameters, several challenges still remain. First, CLaD's compact latent representations may not fully capture visual details for tasks requiring precise manipulation and perceptual ambiguity. CLaD shows reduced performance on Task 9 (put both pots on stove, 81.3\%) compared to other tasks. Object-centric latent representations or spatially-structured foresights that explicitly encode spatial context about objects may help mitigate this problem. Second, our two-stage training framework, which decouples planning from policy optimization, requires approximately 22 hours on a single RTX 4090. Future work should investigate whether Stage 1 dynamics pre-training can be amortized by training on large-scale heterogeneous robot datasets, which may also improve the quality of learned representations across diverse task distributions. Finally, while our approach focuses on manipulation, the principle of learning cross-modal dynamics would generalize to other embodied tasks involving multiple sensory modalities, such as mobile manipulation with vision, force, and tactile feedback, which remains a promising direction for future work.

\clearpage
\setcounter{page}{1}
\maketitlesupplementary
\definecolor{Green1}{rgb}{0.5, 1.0, 0.5} 
\definecolor{Green2}{rgb}{0.65, 1.0, 0.65} 
\definecolor{Green3}{rgb}{0.8, 1.0, 0.8} 

\section{Evaluating Generalizability \& Scalability}
\label{sec:benchmark}

LIBERO benchmark~\cite{liu2023libero} provides three other LIBERO suites except for LIBERO-LONG, each of which contains 10 tasks with 50 demonstrations to evaluate the model's generalizability and scalability over certain properties:

\begin{itemize}
    \item \textbf{LIBERO-Spatial} evaluates the agent’s ability to reason about spatial configurations about diverse layouts of objects;
    \item \textbf{LIBERO-Object} introduces variations in object types while maintaining the scene layouts, assessing the agent’s capacity to generalize across object instances; 
    \item \textbf{LIBERO-Goal} evaluates the generalizability over diverse task objectives given consistent objects and layouts.\vspace{-0.2cm}
\end{itemize}

\begin{table}[h]
    \centering
    \label{tab:libero}
    \caption{\textbf{Average success rates (\%) over 50 rollouts on all suites in LIBERO benchmark.} Color intensity is proportional to the performance level (thicker = higher).}
    \vspace{-0.2cm}
    \renewcommand{\arraystretch}{1.3}
    \begin{tabularx}{\linewidth}{c >{\centering\arraybackslash}X >{\centering\arraybackslash}X >{\centering\arraybackslash}X >{\centering\arraybackslash}X}
    \hline
    \textbf{Method} & \textbf{Spatial} & \textbf{Object} & \textbf{Goal} & \textbf{Long} \\
    \hline
    $\pi_{0}^\diamond$ & 97.2 & \cellcolor{Green3} 97.8 & 91.6 & 82.0 \\
    $\text{UniVLA}$ & 96.5 & 96.8 & \cellcolor{Green3} 95.6 & 92.0 \\
    ${\pi_{0.5}}^\diamond$ & \cellcolor{Green2} 97.6 & \cellcolor{Green2} 98.4 & \cellcolor{Green2} 97.2 & \cellcolor{Green3} 93.2 \\
    $\text{OpenVLA}^\diamond$ & \cellcolor{Green1} 98.2 & \cellcolor{Green1} 98.6 & \cellcolor{Green1} 97.6 & \cellcolor{Green2} 93.8 \\
    \textbf{CLaD (Ours)} & \cellcolor{Green3} 97.3 & 95.7 & 94.3 & \cellcolor{Green1} 94.7 \\
    \hline
    \end{tabularx}
    \raggedright\footnotesize
    $\diamond$: These results are reported in LIBERO-PRO~\cite{liberopro}.
    \vspace{-0.3cm}
\end{table}

CLaD achieves strong performance on long-horizon tasks (i.e., LIBERO-LONG) but sub-optimal results on short-horizon tasks focusing on generalization (i.e., LIBERO-Spatial, Object, Goal), compared to SOTA large VLAs~\cite{kim2024openvla, black2024pi_0, pi0.5, bu2025univla} pre-trained on massive demonstrations, such as Open X-Embodiment~\cite{o2024openx}. Our results indicate that large VLA models leverage substantial pre-trained knowledge for strong in-distribution generalization. CLaD, while having less background knowledge, demonstrates particular strengths in knowledge-independent planning based on cross-modal dynamics modeling. These findings point to CLaD's potential as a building block for scalable robotic planning systems.

\section{Visualization of learned foresight.}

\begin{figure}[t!]
    \centering
    \begin{subfigure}[b]{0.24\linewidth}
        \centering
        \includegraphics[width=\linewidth]{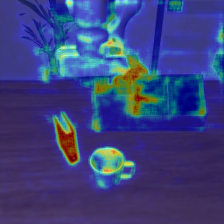}
        \label{fig:sub-a}
    \end{subfigure}
    \begin{subfigure}[b]{0.24\linewidth}
        \centering
        \includegraphics[width=\linewidth]{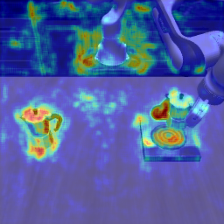}
        \label{fig:sub-b}
    \end{subfigure}
    \begin{subfigure}[b]{0.24\linewidth}
        \centering
        \includegraphics[width=\linewidth]{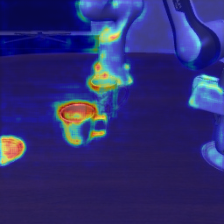}
        \label{fig:sub-c}
    \end{subfigure}
    \begin{subfigure}[b]{0.24\linewidth}
        \centering
        \includegraphics[width=\linewidth]{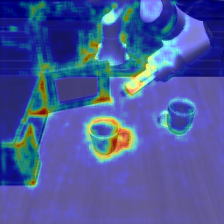}
        \label{fig:sub-d}
    \end{subfigure}
    \vspace{-0.5cm}
    \caption{\textbf{Pixel attribution for predicted latent foresight via Integrated Gradients.} Heatmaps show pixel-level contributions toward the alignment between predicted foresight $\hat{\mathbf{z}}^{t+\tau}$ and target embedding. Brighter regions indicate higher attribution scores. While not yielding precise object boundaries, attributions consistently highlight task-relevant objects, suggesting that the model leverages semantic features for future state prediction.}
    \label{fig:IG}
    \vspace{-0.5cm}
\end{figure}

Figure~\ref{fig:IG} shows which regions influence CLaD's predictions using integrated gradients~\cite{sundararajan2017axiomatic}. The gradient concentrates on task-relevant objects: heatmaps peak near targets in grasping; shift between held object and target in placement. We've confirmed that the predicted foresight attends broadly to object boundaries. While the resulting heatmaps (see Figure. \ref{fig:IG}) do not yield precise object boundaries, they consistently highlight task-relevant objects within the current image. This qualitative analysis confirms that the model relies on meaningful semantic cues to predict future states, demonstrating feasible grounding on observations.

The visualizations show concentrated attention around task-relevant objects: in grasping tasks, heatmaps peak near target objects; in placement tasks, attention shifts between the held object and target location. This suggests CLaD learns spatially grounded predictions rather than relying on global scene context. 

\vspace{-0.2cm}
\section{Effect of Action-Free Data} 
Unlike UVA \cite{bu2025univla}, CLaD's core objective is learning shared dynamics across modalities via asymmetric cross-attention, which fundamentally requires action conditioning. Notably, CLaD already employs stochastic action masking with masking ratio $r=0.3$ to encourage transition inference. To empirically assess action-free data utility, we evaluated Action-free, Heavy Action Mask ($r=0.9$), and Curriculum Learning (initially action-free, then mask with $r=0.3$) variants, but all underperformed the baseline as shown in Table~\ref{tab:action-free}. We hypothesize that removing action guidance introduces multi-modal ambiguity and optimization interference, confirming that our existing strategy is optimal for maintaining deterministic, grounded foresight.

\begin{table}[h]
    \centering
    \setlength{\tabcolsep}{2.3pt}
    \begin{tabular}{l@{\hspace{-1pt}}cccc}
    \toprule
    Variants & Heavy Mask & Action-free & Curriculum & \textbf{CLaD} \\
    \midrule
    Avg. SR (\%) & 88.2 & 90.8 & 85.1 & \textbf{94.7} \\
    \bottomrule
    \end{tabular}
    \caption{Ablation with action-free training variants.}
    \label{tab:action-free}
\end{table}

\vspace{-0.5cm}
\section{Qualitative Results}

\begin{figure*}[t!]
    \centering
    \includegraphics[width=0.75\paperwidth]{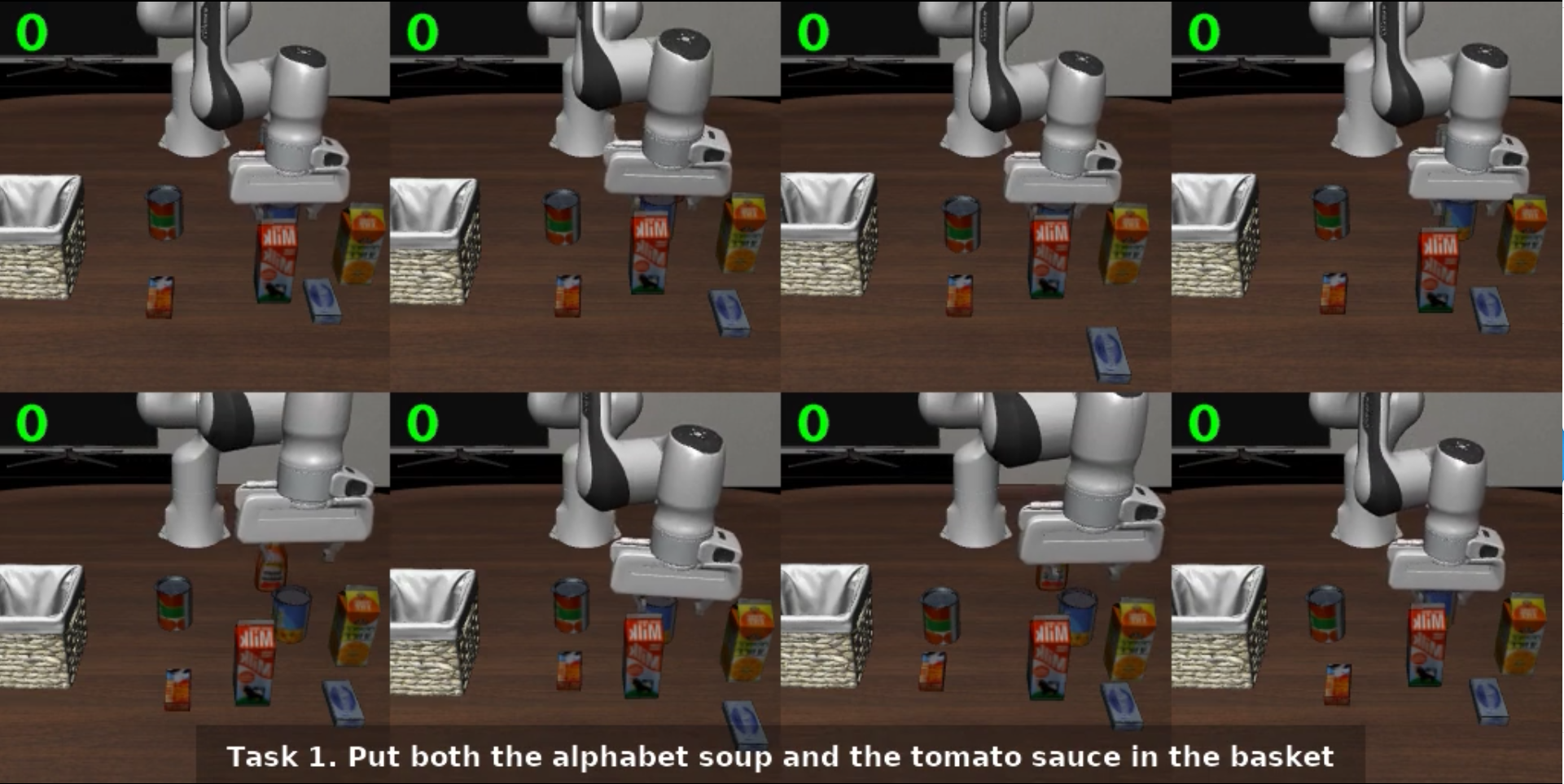}
    \includegraphics[width=0.75\paperwidth]{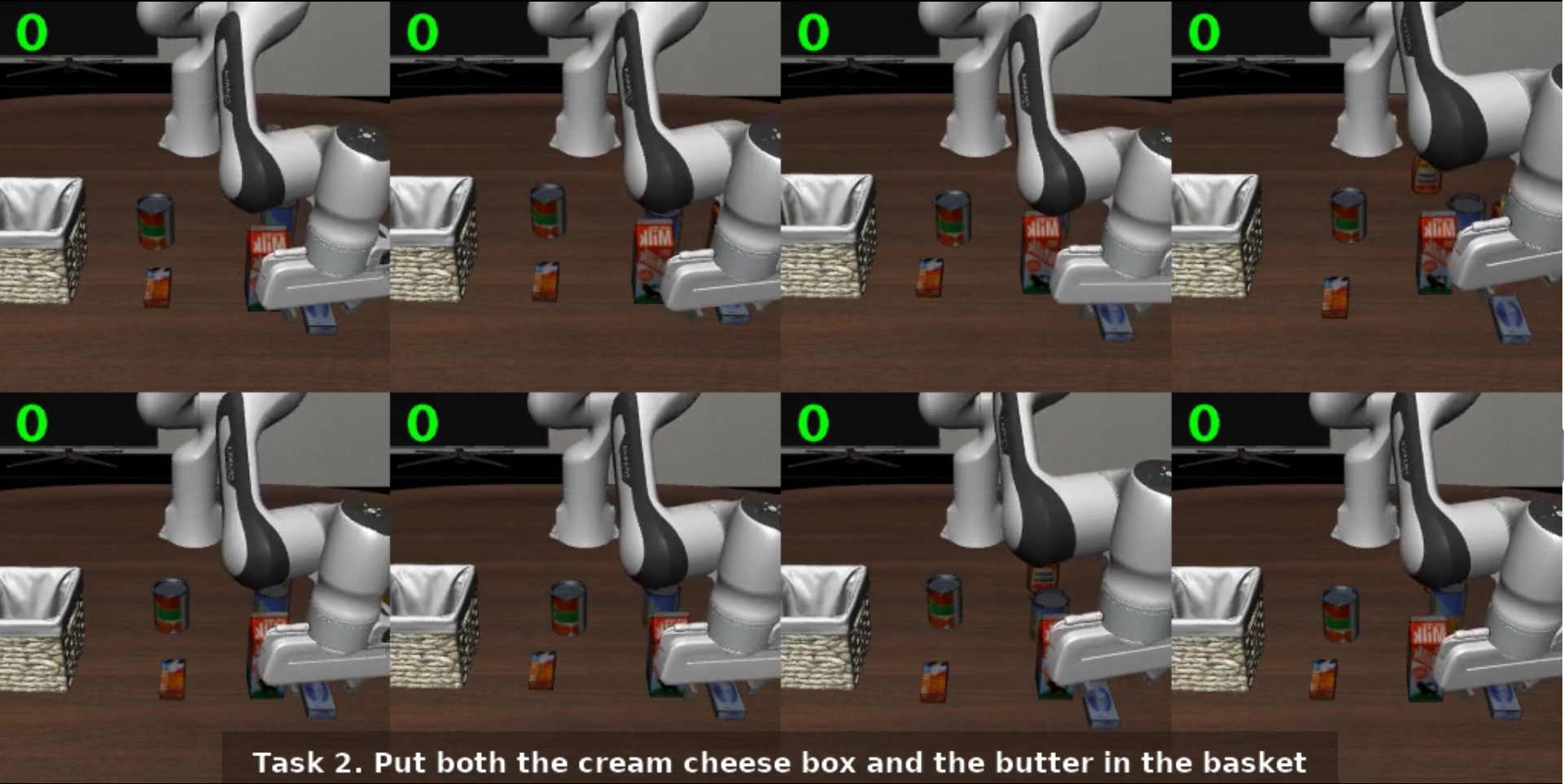}
    \includegraphics[width=0.75\paperwidth]{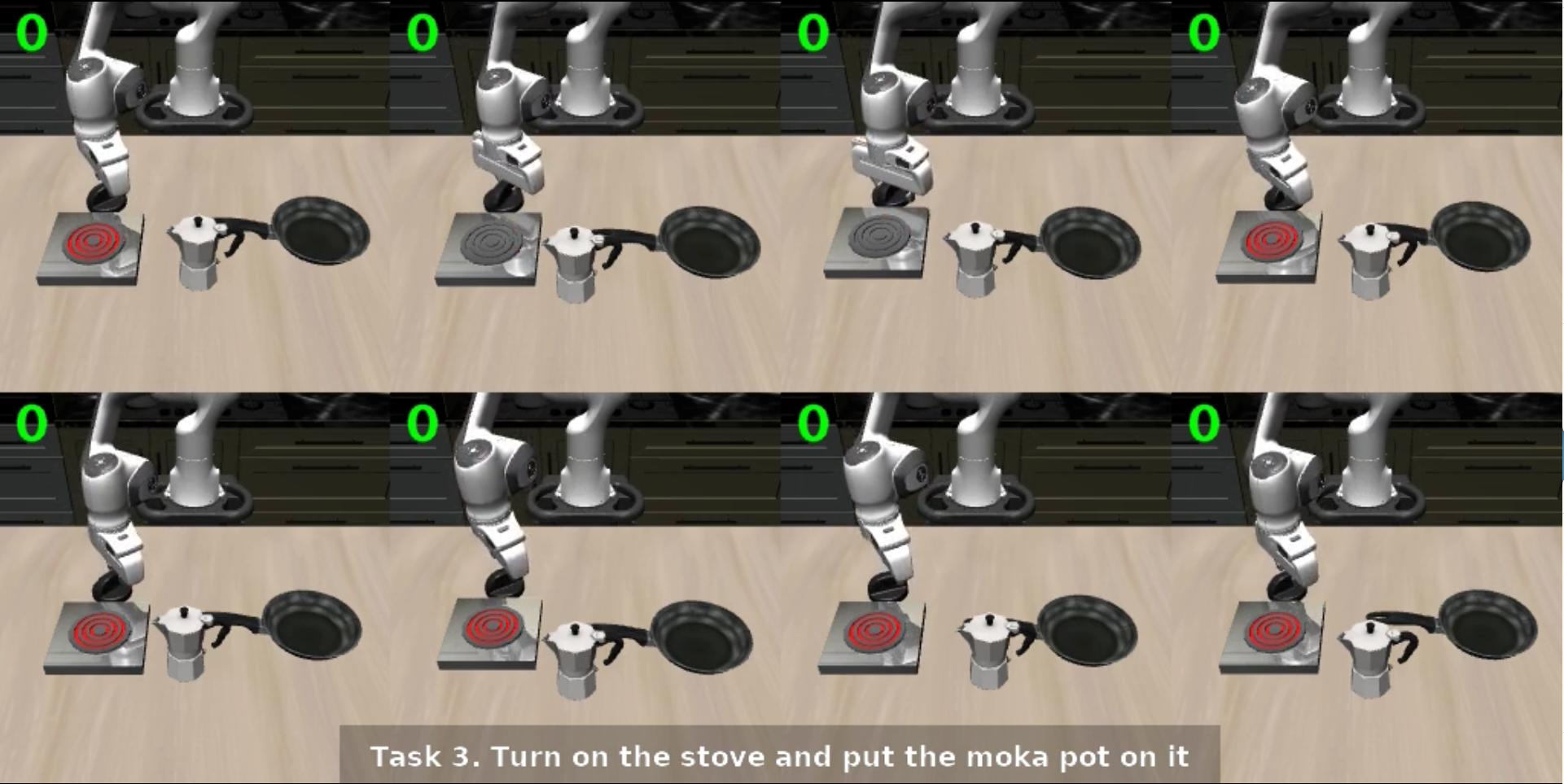}
    \vspace{-0.5em}
    \caption{\textbf{Demonstrations on task 1, 2, 3 of LIBERO-LONG.}}
    \label{fig:libero10-123}
\end{figure*}

\begin{figure*}[!t]
    \centering
    \includegraphics[width=0.75\paperwidth]{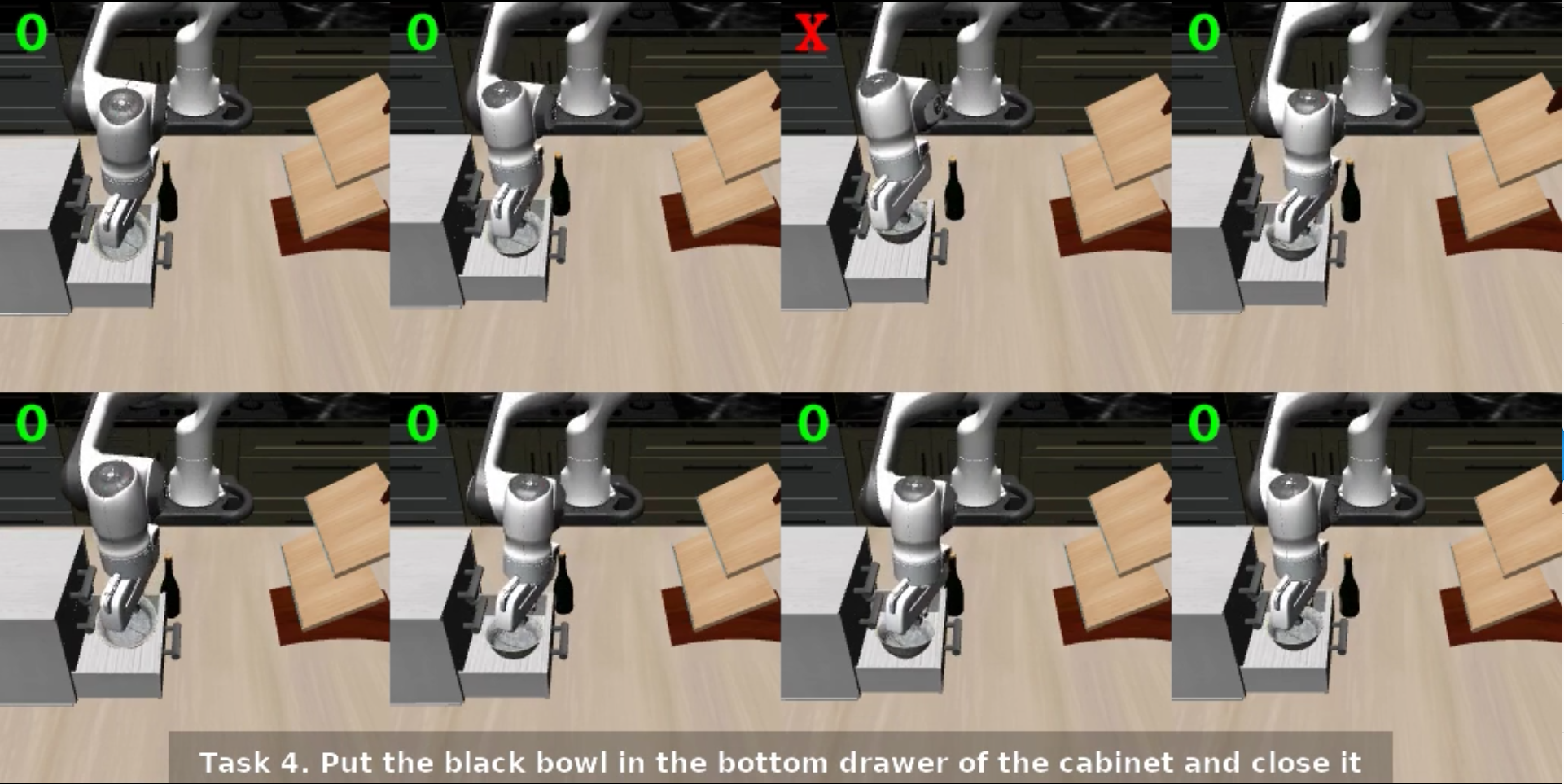}
    \includegraphics[width=0.75\paperwidth]{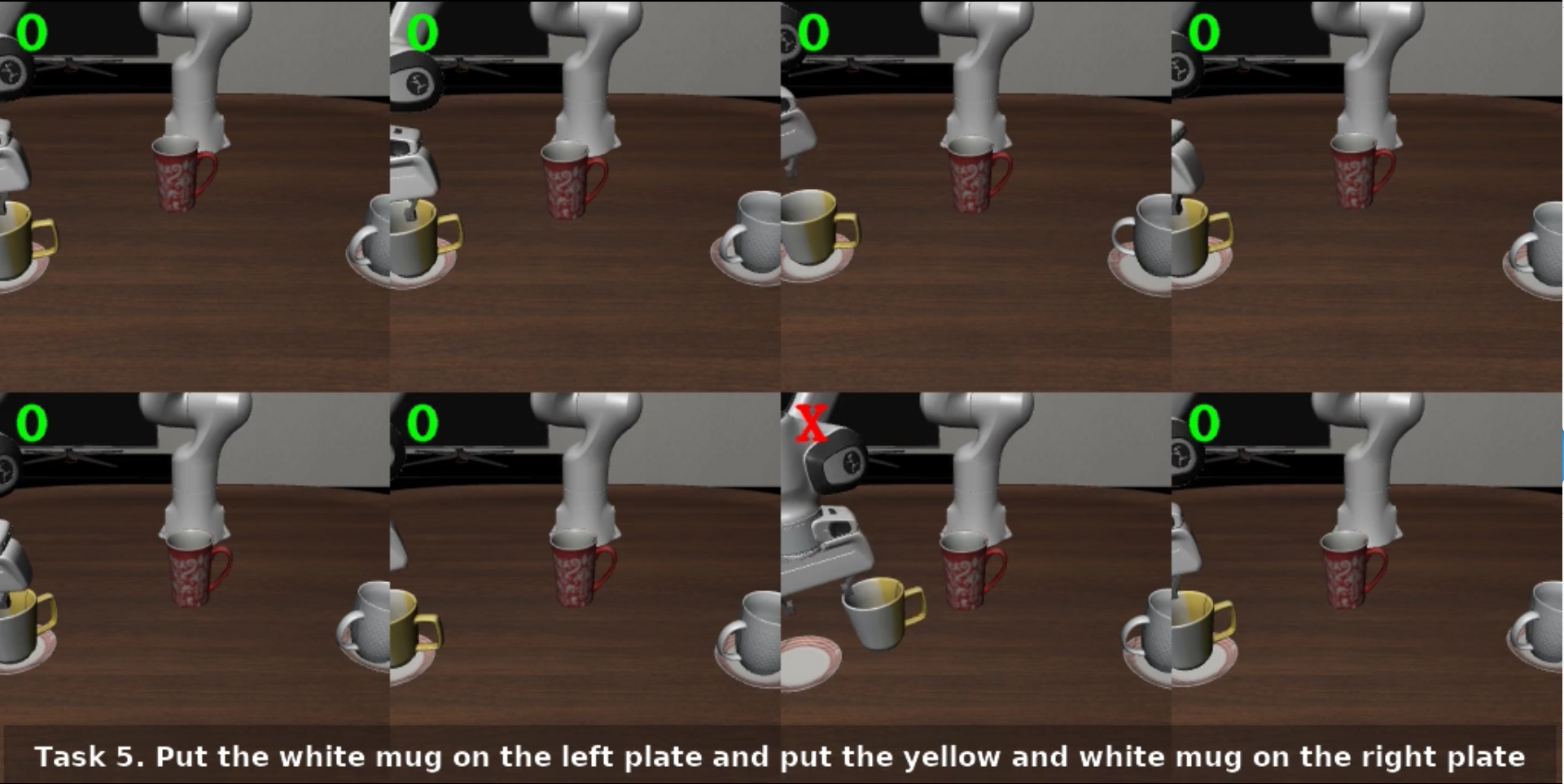}
    \includegraphics[width=0.75\paperwidth]{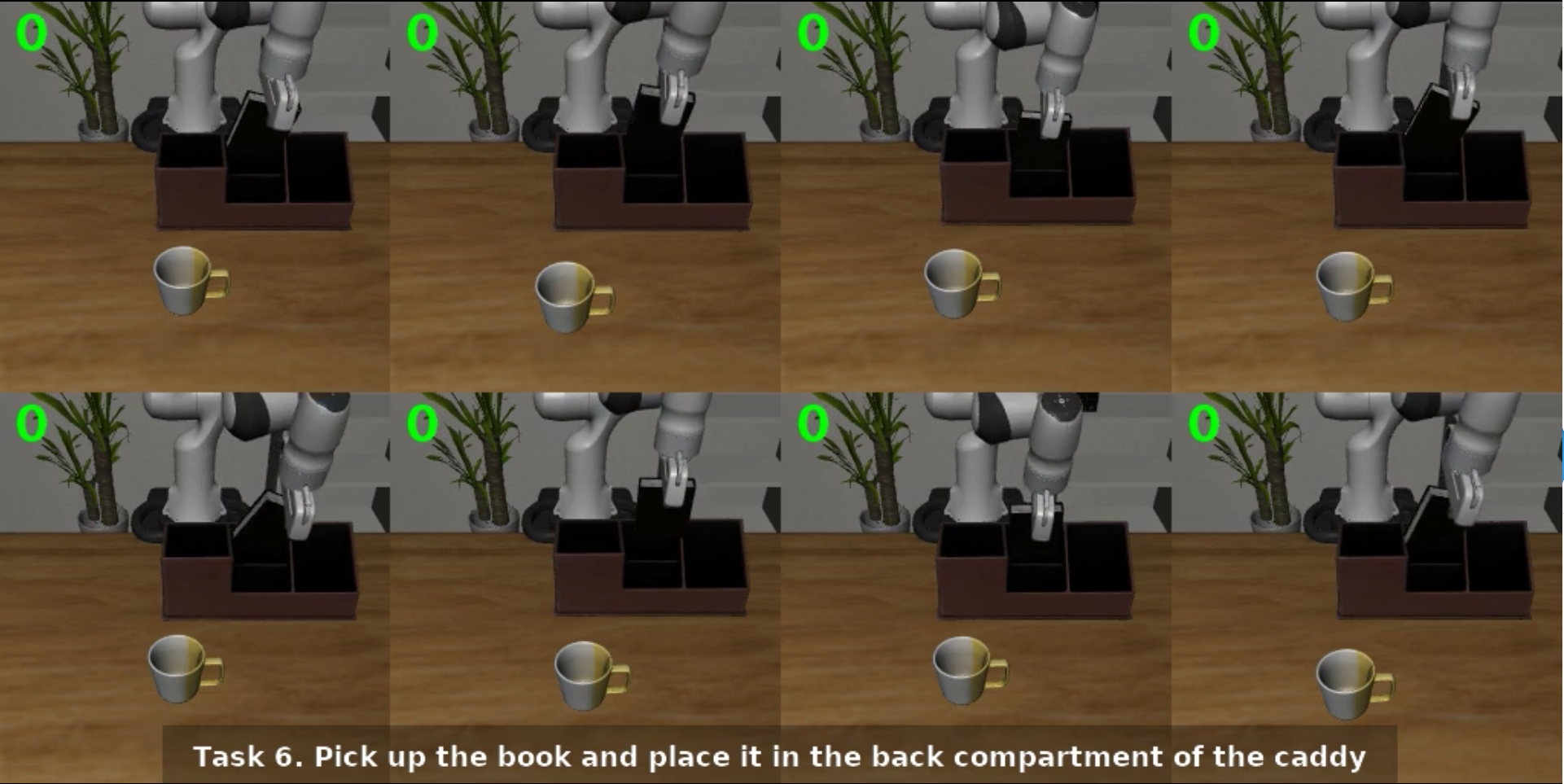}
    \vspace{-0.5em}
    \caption{\textbf{Demonstrations on task 4, 5, 6 of LIBERO-LONG.}}
    \label{fig:libero10-456}
\end{figure*}

\begin{figure*}[!t]
    \centering
    \includegraphics[width=0.75\paperwidth]{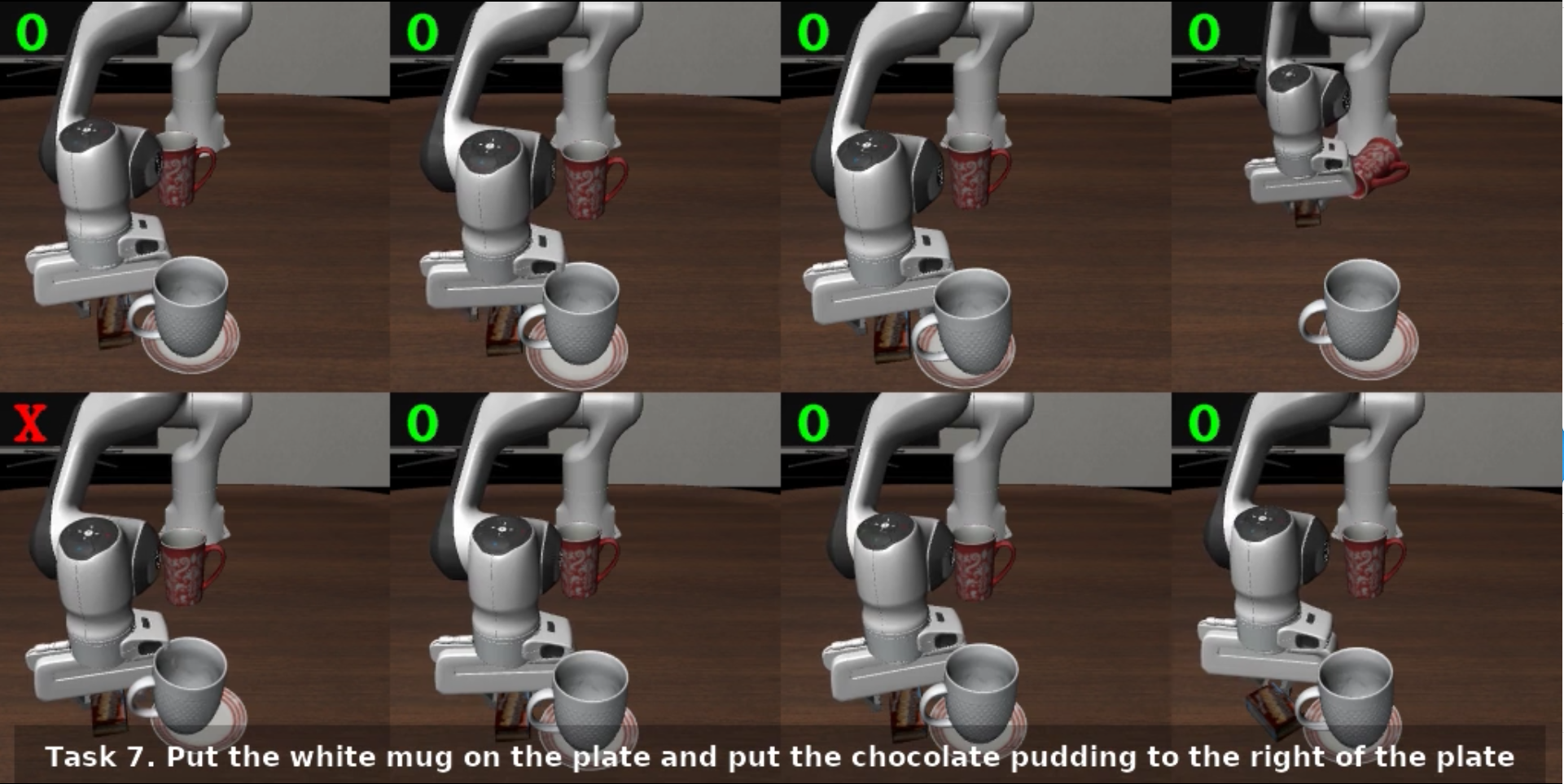}
    \includegraphics[width=0.75\paperwidth]{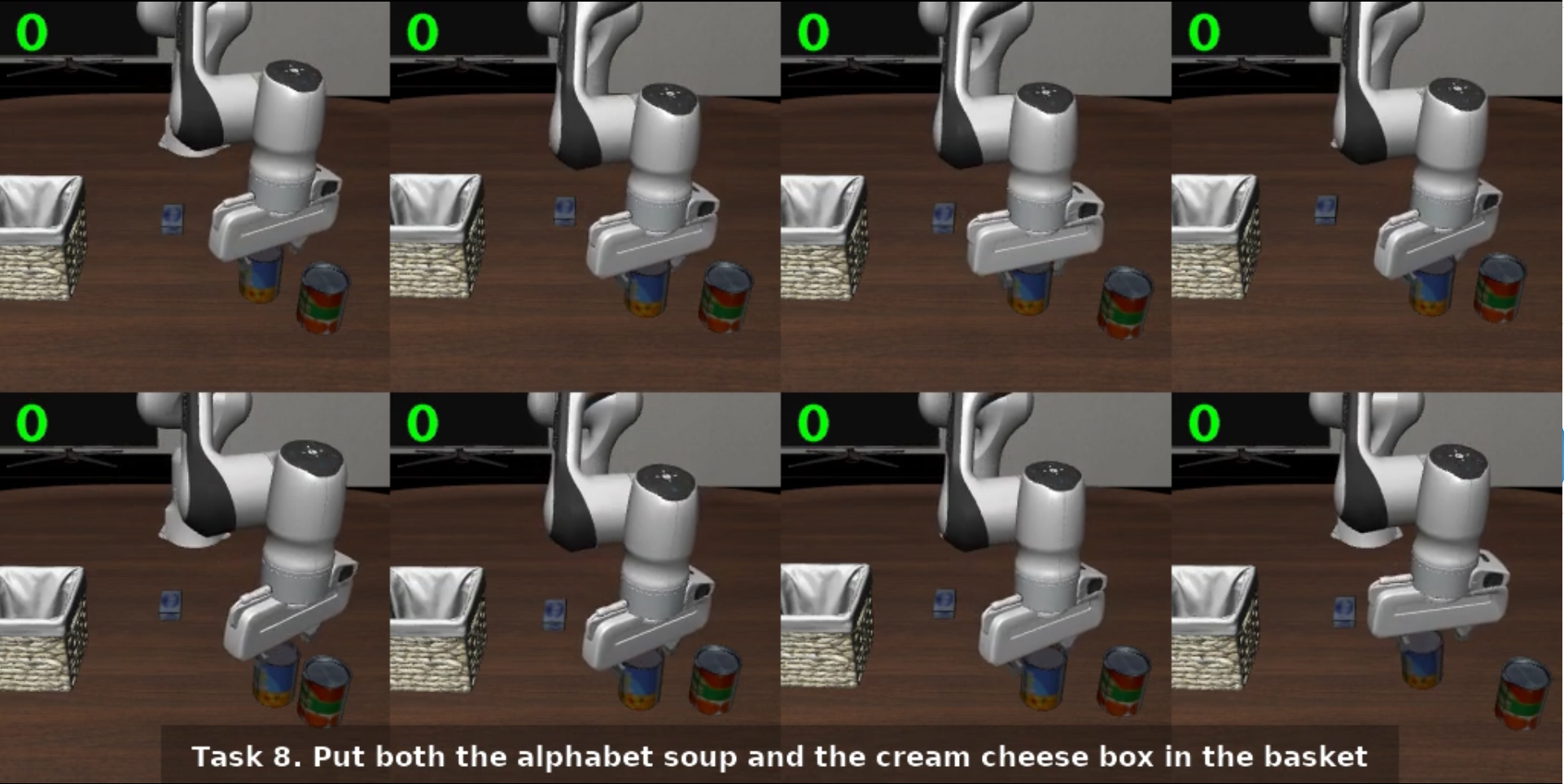}
    \includegraphics[width=0.75\paperwidth]{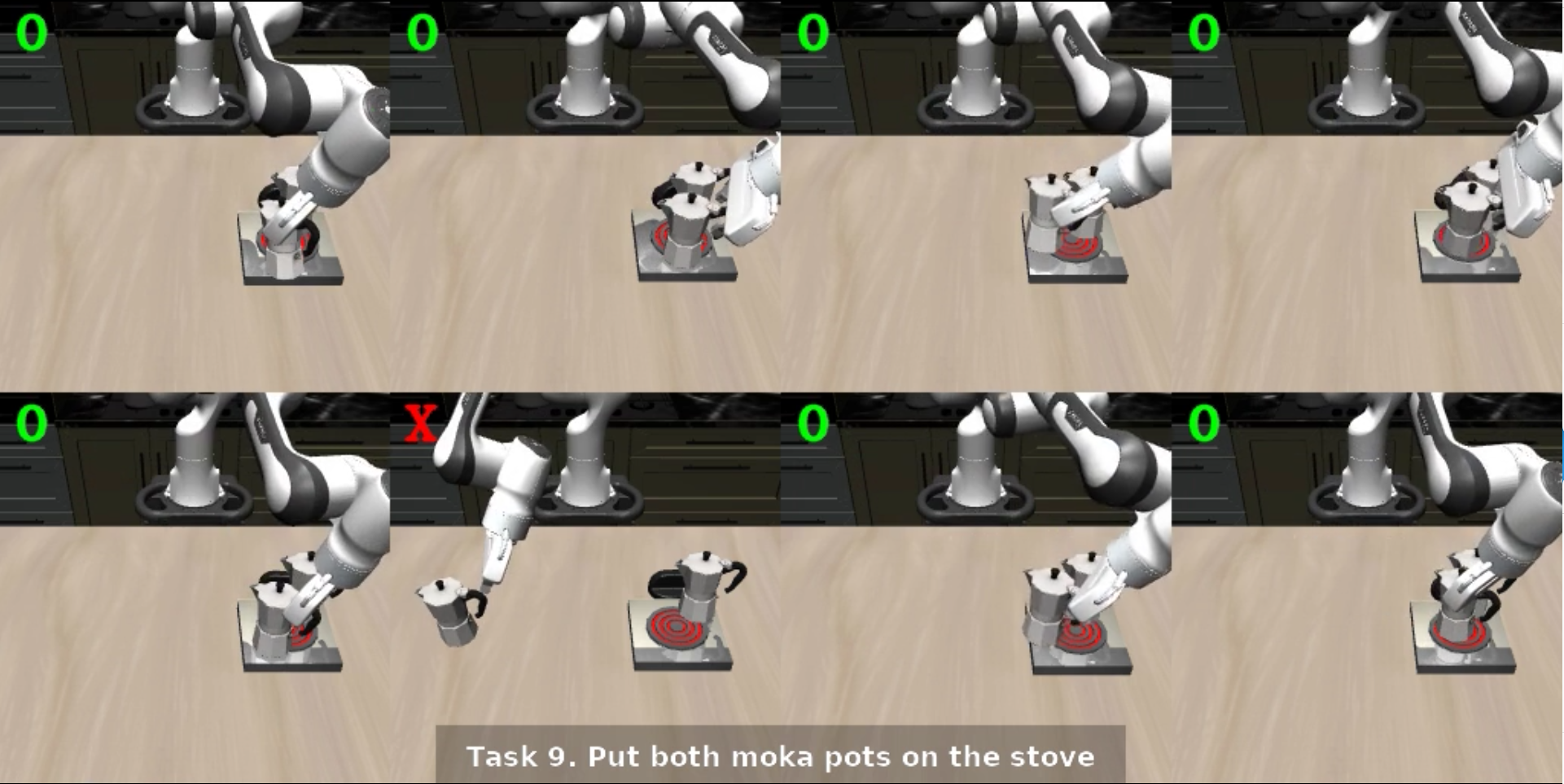}
    \vspace{-0.5em}
    \caption{\textbf{Demonstrations on task 7, 8, 9 of LIBERO-LONG.}}
    \label{fig:libero10-789}
\end{figure*}

\begin{figure*}[!t]
    \centering
    \includegraphics[width=0.75\paperwidth]{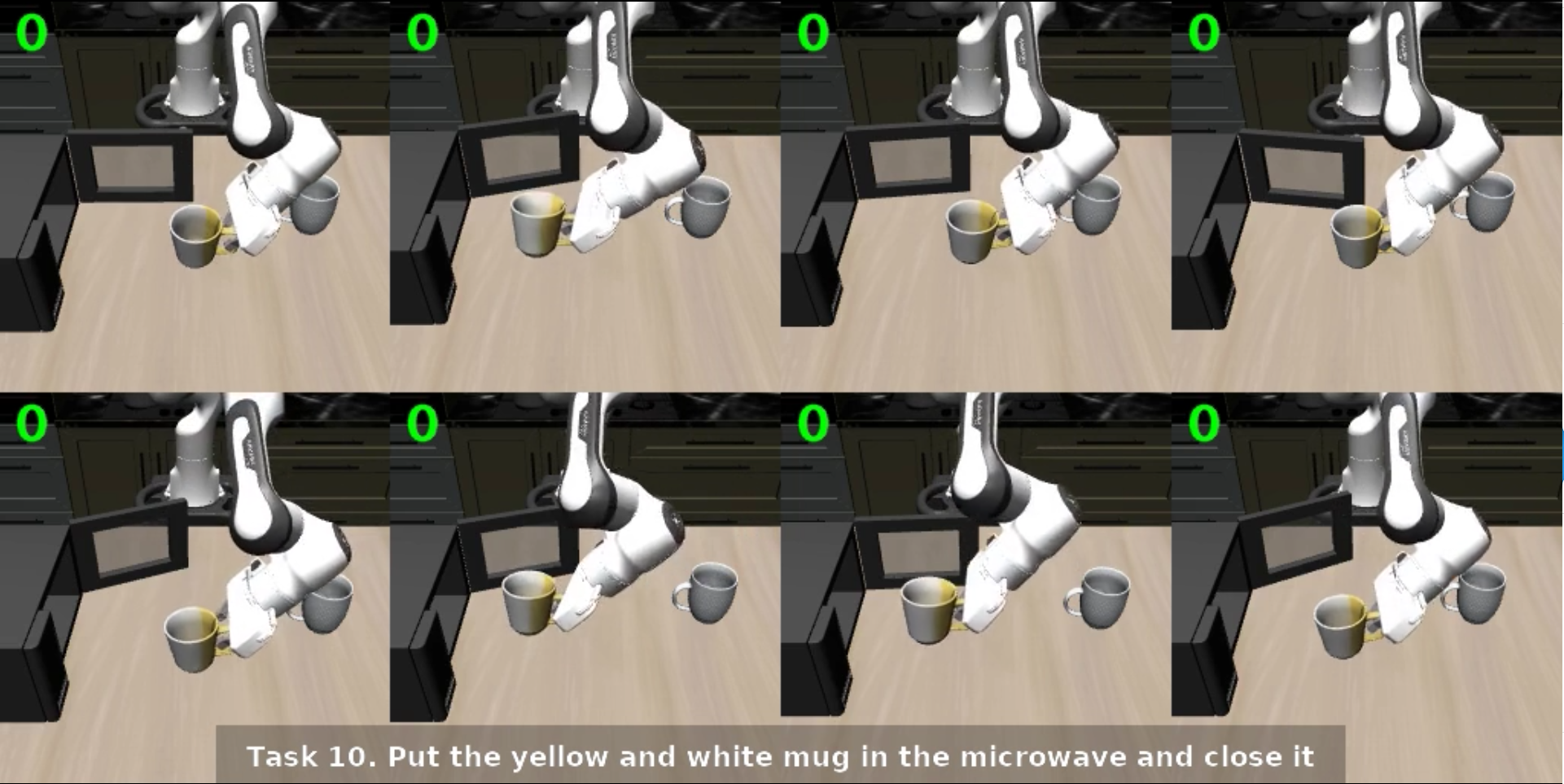}
    \vspace{-0.5em}
    \caption{\textbf{Demonstrations on task 10 of LIBERO-LONG.}}
    \label{fig:libero10-10}
\end{figure*}

\paragraph{Demonstrations on LIBERO-LONG}
Figure. \ref{fig:libero10-123}, \ref{fig:libero10-456}, \ref{fig:libero10-789}, and \ref{fig:libero10-10} visualize representative rollouts for all 10 tasks in LIBERO-LONG.  Green circles mark successful task execution, while red crosses show failures. We provide a supplementary video demonstrating LIBERO-LONG tasks.

\noindent{\textbf{Acknowledgement.}} This work was supported by IITP (Institute of Information \& communications Technology Planning \& Evaluation) and ITRC (Information Technology Research Center) grant funded by the Korea government (MSIT) (RS-2025-25443318, RS-2023-00237965, IITP-2026-RS-2020-II201460). Prof. Sung-Eui Yoon is a corresponding author.
{
    \small
    \bibliographystyle{ieeenat_fullname}
    \bibliography{main}
}

\end{document}